\documentclass{article}

\usepackage{arxiv}

\usepackage[utf8]{inputenc}    
\usepackage[T1]{fontenc}       
\usepackage[colorlinks, linkcolor=black, anchorcolor=black, citecolor=blue]{hyperref} 
\usepackage{url}               
\usepackage{graphicx}          
\usepackage{subcaption}        
\usepackage{booktabs}          
\usepackage{amsmath}           
\usepackage{amssymb}           
\usepackage{amsfonts}          
\usepackage{nicefrac}          
\usepackage{microtype}         
\usepackage{lipsum}            
\usepackage{natbib}            
\usepackage{doi}               
\usepackage{multirow}          
\usepackage{array}             
\usepackage{wrapfig}           
\usepackage{caption}           
\usepackage{tabularx}          
\usepackage{booktabs}          
\usepackage[linesnumbered,ruled,vlined]{algorithm2e} 
\SetKw{KwBy}{by}               
\usepackage{enumitem}          
\usepackage{float}             
\usepackage{makecell}          
\usepackage{bm}                
\usepackage{siunitx}
\usepackage[table]{xcolor}
\title{HiPP-Prune: Hierarchical Preference-Conditioned Structured Pruning for Vision-Language Models}

\date{} 					

\author{
{\large  Lincen BAI$^{1}$ ~ Hedi Tabia$^{1}$ ~ Raúl Santos-Rodríguez$^{2}$} \\
\\
{\large $^{1}$ Université Paris-Saclay   ${^2}$ University of Bristol} 
}

\renewcommand{\headeright}{}
\renewcommand{\undertitle}{}
\renewcommand{\shorttitle}{}

\hypersetup{
pdftitle={HiPP-Prune: Hierarchical Preference-Conditioned Structured Pruning for Vision-Language Models},
pdfsubject={q-bio.NC, q-bio.QM},
pdfauthor={Lincen BAI.~Hedi Tabia, Raúl Santos-Rodríguez},
}

\begin{document}
\maketitle

\begin{abstract}
Pruning vision-language models (VLMs) for efficient deployment is challenging because compression can affect not only task utility but also visual grounding, often amplifying object hallucinations even at the same sparsity level.
We present HiPP-Prune, a hierarchical preference-conditioned structured pruning framework that treats pruning as conditional resource allocation under multiple objectives.
HiPP-Prune makes plan-level decisions: a single policy invocation outputs a global pruning blueprint by factorizing decisions into an overall sparsity budget and a layer-wise allocation, enabling queryable trade-offs via a user-specified preference vector.
To account for VLM-specific failure modes, our policy state integrates a visual sensitivity signal derived from attention flow between vision tokens and language hidden states, discouraging over-pruning of vision-critical layers that facilitate cross-modal fusion.
We optimize pruning plans with plan-level Group Relative Policy Optimization (GRPO) under a multi-objective return that combines task utility, hallucination robustness (POPE), compression, and a synaptic-flow-inspired stability proxy to reduce unproductive exploration in high-sparsity regimes.
Experiments on LLaVA with POPE and ScienceQA demonstrate that HiPP-Prune discovers diverse non-dominated pruning plans and provides controllable robustness--utility trade-offs under matched sparsity budgets.
\end{abstract}


\section{Introduction}

Large vision-language models (VLMs) underpin open-ended multimodal assistants~\cite{liu2023visual}, yet their scale makes deployment costly.
Model pruning is therefore attractive; however, for VLMs \emph{compression alone is not sufficient}.
At the same overall sparsity, pruned models can exhibit markedly different behaviors: conventional task performance may remain stable while hallucination-oriented evaluations degrade.
This is particularly concerning for VLM assistants, where object hallucination---confidently describing entities not supported by the image---is measurable and systematically benchmarked (e.g., POPE~\cite{li2023evaluating}).
These observations motivate treating hallucination robustness as an explicit objective in compression rather than a post-hoc diagnostic, complementing mitigation work that operates at inference time~\cite{leng2024mitigating}.

The above issue reflects a broader challenge in \emph{multi-objective pruning}.
Robustness, utility, and compression often conflict, and the desirable trade-off depends on deployment constraints and risk tolerance.
Moreover, pruning decisions are highly non-uniform across layers~\cite{han2015learning,frankle2018lottery}, and post-training pruning methods for LLMs show that data-dependent criteria can preserve accuracy even at high sparsity~\cite{sun2023simple,frantar2023sparsegpt}.
Together, these results highlight that the central question is not only \emph{how much} to prune, but \emph{where} to allocate sparsity.

Inspired by allocation mechanisms in complex systems, we view VLM pruning as a \emph{conditional resource allocation} problem:
under a fixed compression budget, different preferences over robustness and utility should induce different layer-wise sparsity allocations.
Unlike text-only pruning, VLM pruning must account for components critical to visual grounding.
We therefore incorporate a vision-aware signal---a \emph{visual sensitivity} cue---into the policy state so that layers important for cross-modal fusion can be protected when robustness is prioritized.

Rather than optimizing a single fixed scalarization, HiPP-Prune adopts a \emph{preference-conditioned} formulation:
a single policy adapts to varying preference vectors over robustness, utility, and compression and outputs corresponding pruning plans, approximating a diverse set of Pareto-efficient trade-offs.
This perspective aligns with preference-conditioned multi-objective RL~\cite{roijers2013survey,liu2025efficient} and is practically motivated by modern serving stacks such as vLLM~\cite{kwon2023efficient}, where operating constraints can vary across deployments.

\paragraph{Post-pruning recovery as a probe of structural quality.}
In practice, structured pruning can induce substantial drift on hallucination probes, motivating a lightweight recovery fine-tuning stage that keeps the sparsity pattern fixed.
Using an identical recovery budget across methods provides a controlled lens to compare pruning plans: under matched compression and matched recovery, better plans yield better \emph{pruned initializations} that recover to stronger robustness and utility.

\paragraph{Contributions.}
Our contributions are as follows.
\begin{itemize}
    \item \textbf{Hierarchical preference-conditioned pruning policy.}
    HiPP-Prune casts VLM pruning as conditional resource allocation and learns a hierarchical policy that outputs layer-wise structured sparsity plans by factorizing decisions into global budget control and layer allocation, enabling on-the-fly navigation of the Pareto trade-off space.

    \item \textbf{Attention-flow-based visual sensitivity for vision-aware states.}
    The policy state augments standard layer/activation descriptors with a visual sensitivity cue derived from cross-modal attention flow, highlighting vision-critical components to better preserve visual grounding and hallucination robustness under compression.

    \item \textbf{Plan-level GRPO with SynFlow-gated stabilization.}
    HiPP-Prune trains the plan policy with plan-level GRPO~\cite{mroueh2025revisiting} using margin-based rewards for POPE robustness~\cite{li2023evaluating} and task utility (e.g., ScienceQA~\cite{lu2022learn}), and employs a SynFlow-inspired stability gate to downweight updates from non-viable high-sparsity regimes, stabilizing the combinatorial search.
    Under matched post-pruning recovery budgets, the learned plans yield pruned initializations that recover to stronger robustness--utility trade-offs at the same sparsity.
\end{itemize}

\label{sec:related}

\paragraph{Efficiency and compression for vision-language models.}
The efficiency landscape of vision-language models (VLMs) is shaped by two dominant bottlenecks: the quadratic cost of attention over long multimodal contexts and the parameter/memory footprint of large language backbones.
Accordingly, a substantial body of work pursues \emph{token-/context-level sparsification}, reducing visual tokens or dynamically sparsifying the vision--language context to accelerate attention and KV-cache usage, e.g., instruction-guided or adaptive token pruning~\cite{huang2024ivtp,ye2025atp}, diversity-aware or training-free token pruning~\cite{alvar2025divprune,ye2025fit}, and task- or modality-guided token trimming/pruning for multimodal LLMs~\cite{zhuang2025st3,sun2025lvpruning,huang2025prunevid}.

A complementary line targets \emph{structural} compression, pruning redundant layers or modules to obtain hardware-friendly speedups~\cite{ma2025short,he2025rethinking}.
Learning-based VLM pruning further aims to generalize pruning decisions across prompts and tasks~\cite{liang2025efficientllava}.
Despite this progress, existing approaches largely emphasize either token/context sparsification or fixed structural heuristics, leaving the space of \emph{preference-controllable, layer-wise structured weight allocation} underexplored.
HiPP-Prune addresses this gap by learning a preference-conditioned plan policy that allocates structured sparsity across language-backbone layers while exposing controllable robustness--utility--compression trade-offs.

\paragraph{Post-training pruning for large language models: from criteria to allocation.}
Post-training pruning for large language models (LLMs) has progressed rapidly, with methods such as Wanda~\cite{sun2023simple} and SparseGPT~\cite{frantar2023sparsegpt} showing that data-aware criteria and accurate one-shot updates can preserve performance at high sparsity.
Most methods primarily refine \emph{how to prune}---which weights/rows to remove under a given budget---and reinforce that layer sensitivity is highly non-uniform.
In parallel, structured LLM pruning frameworks (e.g., LLM-Pruner~\cite{ma2023llm}) highlight the practical value of neuron-/channel-level structured sparsity.
Beyond magnitude- or second-order criteria, pruning-at-initialization and signal-propagation analyses (e.g., SynFlow~\cite{tanaka2020pruning}, SNIP~\cite{lee2018snip}, GraSP~\cite{wang2020picking}) emphasize that preserving gradient/flow through the network is critical for avoiding degenerate subnetworks.
For VLMs, an additional challenge is that sparsity distribution can interact non-trivially with cross-modal alignment: sparsifying visually sensitive components can impair grounding even when conventional utility appears stable.
This motivates treating \emph{where to allocate sparsity} as a first-class decision variable under multi-objective constraints.

\paragraph{Post-pruning recovery as a structural integrity probe.}
Structured pruning can induce distribution shift, and lightweight recovery is often adopted to restore behavior while keeping the sparsity pattern fixed.
Parameter-efficient adaptation, such as LoRA~\cite{hu2022lora}, provides a practical mechanism for recovery under a limited trainable-parameter budget.
Prior VLM studies explicitly discuss pruning with restoration strategies~\cite{he2025rethinking}, and LLM work has examined recovery protocols for pruned models under constrained adaptation budgets~\cite{seale2025flexigpt}.
Here, recovery is not framed as ``repairing'' an otherwise unusable pruner; rather, it serves as a controlled probe of structural quality: under matched compression and matched recovery budgets, pruning plans that preserve more structurally favorable subnetworks should recover to stronger robustness--utility trade-offs.

\paragraph{Hallucination robustness: intrinsic versus extrinsic interventions.}
Object hallucination in VLMs has been systematically characterized and benchmarked by POPE~\cite{li2023evaluating}, enabling controlled evaluation of groundedness.
Beyond POPE, diagnostic suites such as HallusionBench further stress-test entangled hallucination and visual illusion failure modes in large VLMs~\cite{guan2024hallusionbench}.
Mitigation methods such as Visual Contrastive Decoding (VCD)~\cite{leng2024mitigating} represent \emph{extrinsic} interventions that adjust inference-time behavior without changing the underlying model structure.
In contrast, robustness under compression introduces an \emph{intrinsic} dimension: the sparsified architecture itself can amplify or suppress hallucination tendencies depending on how sparsity is distributed across vision-relevant pathways.
This perspective motivates treating hallucination robustness as an explicit consideration during pruning, rather than solely as a post-hoc diagnostic.

\paragraph{Plan-level decision making under multi-objective preferences.}
Multi-objective reinforcement learning (MORL) seeks Pareto-efficient solutions across varying preferences~\cite{roijers2013survey}, with recent focus on efficient discovery of the Pareto front at test time~\cite{liu2025efficient}.
HiPP-Prune instantiates this paradigm by casting pruning as a one-shot, plan-level allocation, where a single policy decision specifies a global layer-wise sparsity blueprint rather than a sequence of incremental actions.
While Group Relative Policy Optimization (GRPO) typically targets token-level alignment~\cite{mroueh2025revisiting}, we extend its advantage mechanism to the combinatorial space of pruning plans by comparing candidate blueprints under a fixed preference $\mathbf{w}$.
This internalizes robustness--utility--compression trade-offs into the structural blueprint, enabling zero-shot querying of operating points without retraining specialized pruners.
Furthermore, while stability in policy optimization is often achieved via trust-region or constrained methods~\cite{pmlr-v37-schulman15,schulman2017proximal,pmlr-v70-achiam17a}, HiPP-Prune employs a SynFlow-inspired signal as a compute-aware stability gate to filter non-viable pruning episodes during high-sparsity search.

\section{Method}
\label{sec:method}

\subsection{Preliminaries}
\label{sec:preliminaries}

Let $f_{\theta_0}$ denote a pretrained vision-language model (VLM), e.g., LLaVA~\cite{liu2023visual}, consisting of a vision encoder and a language backbone.
We consider \emph{structured} pruning of the language backbone by applying row-wise (neuron-wise) masks to a set of linear projections indexed by $\mathcal{L}$ with $|\mathcal{L}|=L$.

\paragraph{Pruning plan.}
A pruning plan is a layer-wise ratio vector $\mathbf{r}=[r_1,\dots,r_L]^\top\in[0,1]^L$, where $r_\ell$ specifies the pruning ratio (fraction removed) for layer $\ell$.
Applying a plan produces a pruned model
\begin{equation}
    \theta(\mathbf{r})=\mathcal{P}(\theta_0,\mathbf{r}),
\end{equation}
where $\mathcal{P}(\cdot)$ is our structured pruning operator.

\paragraph{Multi-objective returns, preferences, and stability regularization.}
We evaluate a plan $\mathbf{r}$ with a three-dimensional objective vector
\begin{equation}
\label{eq:objective_vector}
    \mathbf{J}(\mathbf{r})=
    \big[J_{\text{rob}}(\mathbf{r}),\,J_{\text{util}}(\mathbf{r}),\,J_{\text{comp}}(\mathbf{r})\big]^\top,
\end{equation}
corresponding to hallucination robustness, task utility, and compression, respectively.
Following preference-conditioned multi-objective learning~\cite{roijers2013survey,liu2025efficient}, we sample a preference vector $\mathbf{w}\in\Delta^{2}$ to represent deployment-specific trade-offs and optimize the scalarized objective
\begin{equation}
\label{eq:scalarized_objective}
    J_{\mathbf{w}}(\mathbf{r})=\mathbf{w}^\top\,\mathrm{norm}\!\left(\mathbf{J}(\mathbf{r})\right),
\end{equation}
where $\mathrm{norm}(\cdot)$ denotes online normalization to mitigate scale imbalance across objectives.
In addition, a synaptic-flow-inspired stability signal is used as a \emph{regularization penalty} during exploration to prune away non-viable network topologies under high sparsity, \emph{decoupled} from the user-specified trade-off vector $\mathbf{w}$.

\paragraph{Plan-level policy.}
We learn a preference-conditioned policy that outputs a \emph{global} pruning blueprint in one-shot:
\begin{equation}
\label{eq:plan_policy}
    \mathbf{a}\sim\pi_\phi(\mathbf{a}\mid \mathbf{s},\mathbf{w}), \qquad \mathbf{r}=g(\mathbf{a}),
\end{equation}
where $\mathbf{a}$ is the raw action and $g(\cdot)$ maps actions to a valid layer-wise plan by enforcing budget and box constraints (i.e., $\mathbf{r}\in[0,1]^L$ and overall sparsity within a target range).
The state representation $\mathbf{s}$ encapsulates layer-wise activation statistics and visual sensitivity signals to facilitate cross-modal grounding.
Unlike token-level RL used in dialogue alignment, our decision is plan-level: one policy invocation specifies the entire sparsity allocation $\mathbf{r}$, which is then evaluated by $\mathbf{J}(\mathbf{r})$.

\subsection{HiPP-Prune Overview}
\label{sec:hipp_overview}

\begin{figure*}[t]
    \centering
    \includegraphics[width=0.96\textwidth]{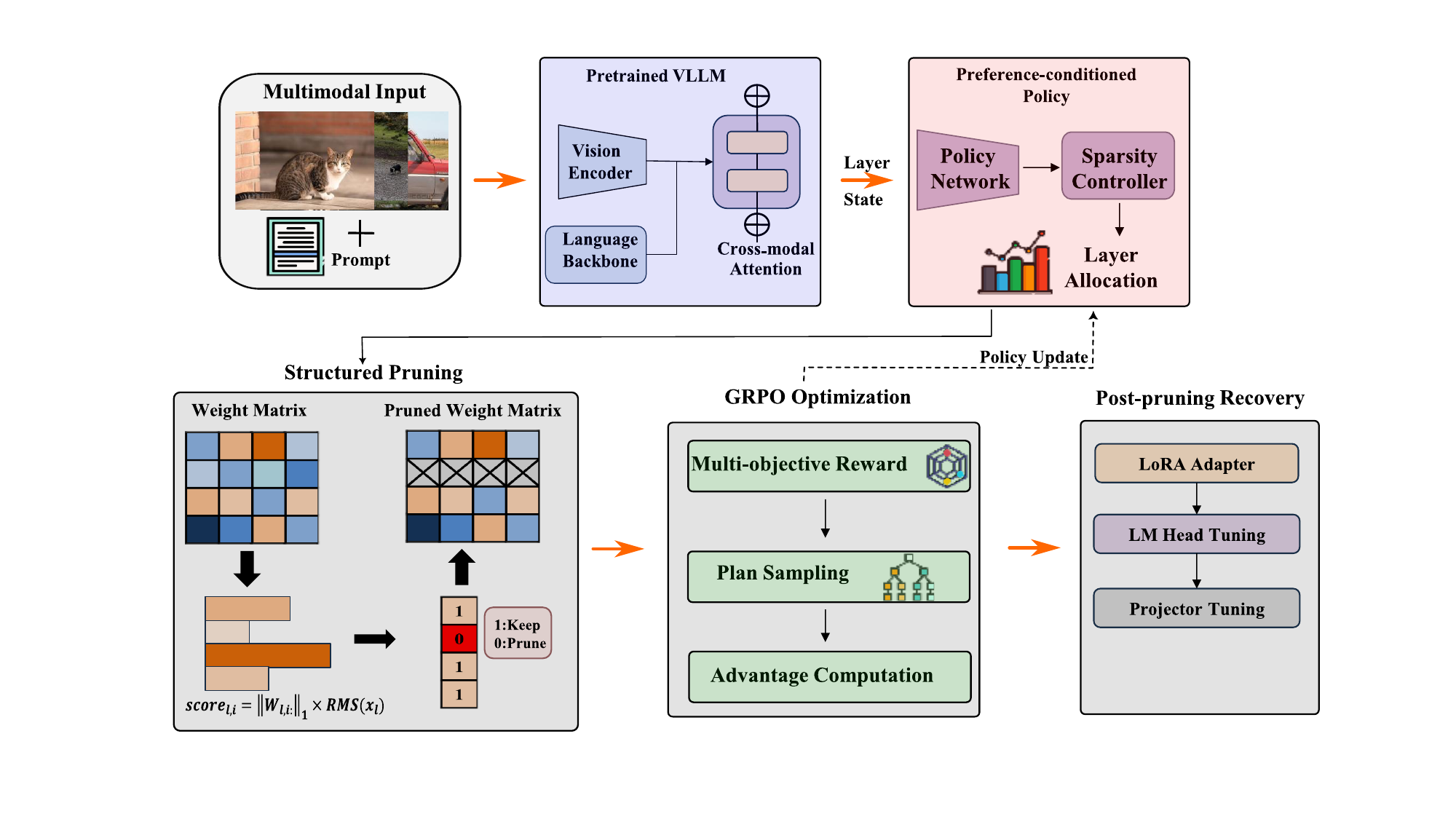}
    \caption{\textbf{An overview of HiPP-Prune.}
    A pretrained VLM and calibration data are used to construct layer-wise states, integrating activation statistics with vision-aware cross-modal cues.
    Conditioned on a preference vector, a hierarchical policy generates a one-shot structured pruning plan by factorizing decisions into global sparsity control and layer-wise allocation.
    Candidate plans are optimized via plan-level GRPO using robustness, utility, and compression feedback, while a SynFlow-inspired stability gate downweights updates from non-viable high-sparsity episodes to stabilize plan search.
    Finally, the optimized pruning plan is applied to the language backbone, followed by lightweight post-pruning recovery fine-tuning.}
    \label{fig:overview_pipeline}
\end{figure*}

HiPP-Prune (\textbf{Hi}erarchical \textbf{P}reference-conditioned \textbf{P}runing) is a plan-level framework for allocating structured sparsity in VLMs under the competing objectives of robustness, utility, and compression.
As illustrated in Fig.~\ref{fig:overview_pipeline}, a sampled preference vector $\mathbf{w}$ first conditions a vision-aware layer-state representation, whose features combine standard pruning cues with cross-modal sensitivity signals that highlight components important for visual grounding (Sec.~\ref{sec:state_repr}).
A hierarchical policy then generates a global pruning plan $\mathbf{r}$ in a one-shot manner by factorizing decisions into budget control and layer-wise allocation, where $g(\cdot)$ maps the factorized action to a valid plan (Sec.~\ref{sec:hier_policy}).
The resulting plan is applied by a structured pruning operator $\mathcal{P}$ and evaluated using robustness on POPE~\cite{li2023evaluating}, utility on ScienceQA~\cite{lu2022learn}, and compression feedback.
To stabilize exploration in high-sparsity regimes, HiPP-Prune further employs a SynFlow-inspired stability gate~\cite{tanaka2020pruning} that downweights policy updates from non-viable pruning episodes rather than treating stability as a user-facing objective (Sec.~\ref{sec:objectives_grpo}).
The policy is finally optimized with plan-level GRPO~\cite{mroueh2025revisiting}, using group-relative advantages to efficiently discover diverse Pareto-efficient pruning strategies.

\subsection{Vision-aware State Representation}
\label{sec:state_repr}

\paragraph{Layer-wise state.}
For each prunable layer $\ell\in\mathcal{L}$, the policy consumes a state vector $\mathbf{s}_\ell$ composed of layer identity/type, weight and activation statistics estimated on a calibration set, the current budget context, the preference vector $\mathbf{w}$, and a vision-aware cue termed \emph{visual sensitivity}. The state concatenates a fixed set of layer descriptors with the preference vector.

\paragraph{Visual sensitivity via cross-modal attention mass.}
VLM pruning can disproportionately harm visual grounding; therefore, the policy state includes a scalar \emph{visual sensitivity} score to highlight layers that facilitate cross-modal fusion.
Let $\mathcal{V}$ denote the set of vision tokens and $\mathcal{T}$ the set of language tokens for an input sequence.
For transformer block $\ell$, let $\mathbf{A}^{(\ell,h)}\in\mathbb{R}^{|\mathcal{T}|\times|\mathcal{V}|}$ denote the (softmax-normalized) attention from language tokens to vision tokens for head $h\in\{1,\dots,H\}$, and define the head-averaged attention mass
\begin{equation}
\label{eq:head_agg}
\bar{\mathbf{A}}^{(\ell)} \;=\; \frac{1}{H}\sum_{h=1}^{H}\mathbf{A}^{(\ell,h)} .
\end{equation}
We use a layer-wise proxy for cross-modal reliance,
\begin{equation}
\label{eq:vis_sens}
\mathcal{S}_\ell \;=\; \max_{t\in\mathcal{T}} \sum_{v\in\mathcal{V}} \bar{\mathbf{A}}^{(\ell)}_{t,v},
\end{equation}
and include $\mathcal{S}_\ell$ as the visual sensitivity feature in $\mathbf{s}_\ell$.
Using the maximum rather than the mean prevents the dilution of sparse but critical cross-modal grounding signals across the entire language sequence.
Here $\bar{\mathbf{A}}^{(\ell)}$ is head-averaged; alternative aggregations (e.g., sum or max over heads) yield similar behavior in practice.

\paragraph{Computation and overhead.}
To avoid additional RL training overhead, visual sensitivity is computed once prior to policy training using a fixed, small calibration batch (the same type of data used for activation statistics), and is then treated as a static layer-wise cue throughout training.
For distributed training, the resulting per-layer sensitivity scores are synchronized across workers to ensure consistent state features.

\subsection{Plan Parameterization and Hierarchical Policy}
\label{sec:hier_policy}

Directly predicting an independent pruning ratio for each layer yields a high-dimensional continuous action space, which is unstable for plan search and makes it difficult to enforce a global sparsity budget.
HiPP-Prune instead adopts a \emph{hierarchical} plan parameterization that separates \emph{budget control} from \emph{layer allocation}.
Let $\mathbf{S}=\{\mathbf{s}_\ell\}_{\ell\in\mathcal{L}}$ denote the collection of layer states.
The action is $\mathbf{a}=\{s,\mathbf{p}\}$, where $s\in(0,1)$ is a scalar global sparsity controller and $\mathbf{p}\in\Delta^{L-1}$ is a simplex-valued allocation vector:
\begin{equation}
\label{eq:beta_dirichlet}
\begin{aligned}
s &\sim \mathrm{Beta}\!\big(\alpha_\phi(\mathbf{S},\mathbf{w}),\,\beta_\phi(\mathbf{S},\mathbf{w})\big),\\
\mathbf{p} &\sim \mathrm{Dirichlet}\!\big(\boldsymbol{\eta}_\phi(\mathbf{S},\mathbf{w})\big).
\end{aligned}
\end{equation}
Although $s$ and $\mathbf{p}$ are sampled from separate heads, they share the same feature backbone conditioned on $(\mathbf{S},\mathbf{w})$, which couples budget control and allocation through shared representations; the current implementation does not additionally condition $\mathbf{p}$ on the sampled $s$.
We map $s$ to a target sparsity interval $[c_{\min},c_{\max}]$ and combine $(s,\mathbf{p})$ into layer-wise ratios:
\begin{equation}
\label{eq:plan_mapping}
    \tilde{s} = c_{\min} + s(c_{\max}-c_{\min}), \qquad
    r_\ell = \mathrm{clip}\!\left(\kappa \,\tilde{s}\, L\, p_\ell,\; 0,\; 1\right).
\end{equation}
Here $\kappa$ controls the \emph{concentration} of layer-wise allocation induced by $\mathbf{p}$: larger $\kappa$ encourages sharper, more uneven allocations, while smaller $\kappa$ yields smoother distributions.
Because Eq.~\eqref{eq:plan_mapping} applies $\mathrm{clip}(\cdot)$, excessively large $\kappa$ may saturate some layers at $r_\ell=1$, which can cause the realized overall sparsity to deviate from the target implied by $\tilde{s}$.
In practice, $\kappa$ is set to $1.0$ by default and tuned to avoid extensive saturation; the resulting realized sparsity remains within the target budget range $[c_{\min},c_{\max}]$.
In our experiments, $[c_{\min},c_{\max}]$ typically spans moderate-to-high compression (e.g., $[0.2,0.5]$), enabling controllable operating points within a fixed deployment budget.
This factorization reduces search complexity and naturally enforces a global sparsity budget.

\subsection{Structured Pruning Operator}
\label{sec:pruning_operator}

\paragraph{Role of the operator.}
The pruning operator $\mathcal{P}$ executes a layer-wise plan $\mathbf{r}$ by applying structured masks, yielding the pruned model $\theta(\mathbf{r})=\mathcal{P}(\theta_0,\mathbf{r})$.
Its \emph{vision-aware} behavior does not come from the operator itself, but from the policy: visual sensitivity shapes the ratios $r_\ell$, which determine how aggressively each layer is pruned.

\paragraph{Structured row-wise masking.}
Given plan ratios $\mathbf{r}$, we apply structured row-wise masking to each prunable linear projection.
In the language backbone, this is performed on MLP projections (e.g., gate/up/down), where each mask entry corresponds to an output neuron (output channel).
For paired projections such as gate and up in LLaMA/Vicuna-style MLPs, we enforce ratio-level consistency within the same block.
For a weight matrix $\mathbf{W}_\ell\in\mathbb{R}^{d_{\text{out}}\times d_{\text{in}}}$, we maintain a binary mask $\mathbf{m}_\ell\in\{0,1\}^{d_{\text{out}}}$ and prune a fraction $r_\ell$ of rows by setting entries in $\mathbf{m}_\ell$ to zero.
The masked weight is
\begin{equation}
    \hat{\mathbf{W}}_\ell = \mathrm{diag}(\mathbf{m}_\ell)\mathbf{W}_\ell.
\end{equation}
This row-wise design yields hardware-friendly structured sparsity and is compatible with efficient implementations when the masked structure is materialized.

\paragraph{Data-aware importance and calibration.}
To select rows, we adopt a Wanda-style criterion~\cite{sun2023simple} that combines weight magnitude with calibration-time activation statistics.
Let $\mathrm{RMS}(\mathbf{x}_\ell)$ denote a summary of the input activation magnitude for layer.
Each output row $i$ is scored as
\begin{equation}
\label{eq:wanda_score}
    \mathrm{score}_{\ell,i} = \|\mathbf{W}_{\ell,i:}\|_1 \cdot \mathrm{RMS}(\mathbf{x}_\ell),
\end{equation}
and the lowest-scoring fraction $r_\ell$ is pruned.

\subsection{Multi-objective Returns and Plan-level GRPO}
\label{sec:objectives_grpo}

\paragraph{Robustness, utility, and compression.}
We define robustness and utility returns using smooth log-margin objectives, which provide denser learning signals than discrete accuracy.
For an instance, let $\mathcal{Y}_{\text{gt}}$ denote the set of target answer tokens (e.g., \{Yes, No\} for POPE~\cite{li2023evaluating}) and $\mathcal{Y}_{\text{neg}}$ denote a set of competing tokens.
We compute
\begin{equation}
\label{eq:log_margin}
    J_{\text{margin}} = \log \sum_{y\in\mathcal{Y}_{\text{gt}}} P_\theta(y) \;-\; \log \sum_{y\in\mathcal{Y}_{\text{neg}}} P_\theta(y),
\end{equation}
and instantiate it for $J_{\text{rob}}$ and $J_{\text{util}}$ on the corresponding benchmarks.
Compression return $J_{\text{comp}}$ is computed from the realized overall sparsity and encourages meeting a deployment budget: it increases linearly from $0$ to $1$ as sparsity moves from $c_{\min}$ to $c_{\max}$, and saturates outside $[c_{\min},c_{\max}]$.

\paragraph{Running normalization (online update).}
To align scales across objectives, $\mathrm{norm}(\cdot)$ uses a running mean and standard deviation updated online from the current group's raw scores, followed by clipping for stability (we clip normalized values to $[-5,5]$).
We denote
\begin{equation}
\label{eq:norm_defs}
n_{\text{rob}}(\mathbf{r})=\mathrm{norm}\!\left(J_{\text{rob}}(\mathbf{r})\right),\qquad
n_{\text{util}}(\mathbf{r})=\mathrm{norm}\!\left(J_{\text{util}}(\mathbf{r})\right).
\end{equation}

\paragraph{SynFlow-inspired stability gating.}
High-sparsity exploration may produce non-viable pruning topologies.
Following SynFlow~\cite{tanaka2020pruning}, we compute an episode-level stability signal as the log-ratio
\begin{equation}
\label{eq:syn_logratio}
    \rho_{\text{syn}}(\mathbf{r}) \;=\; \log\frac{\mathrm{Flow}(\theta(\mathbf{r}))+\epsilon}{\mathrm{Flow}(\theta_0)+\epsilon},
\end{equation}
evaluated once per episode on a single worker and synchronized across workers for efficiency.
A two-sided acceptable band $[\ell_t,u_t]$ is maintained via quantile-based calibration and warmup annealing, and deviations are penalized by a band loss
\begin{equation}
\label{eq:syn_shaping}
\psi(\rho_{\text{syn}};\ell_t,u_t)=
\begin{cases}
-\lambda_{\text{syn}}(\ell_t-\rho_{\text{syn}}), & \rho_{\text{syn}}<\ell_t,\\
0, & \ell_t \le \rho_{\text{syn}}\le u_t,\\
-\lambda_{\text{syn}}(\rho_{\text{syn}}-u_t), & \rho_{\text{syn}}>u_t,
\end{cases}
\end{equation}
where $\lambda_{\text{syn}}>0$ controls the penalty slope outside the band (we use $\lambda_{\text{syn}}=2$ by default).
We convert this shaping signal into an episode importance weight
\begin{equation}
\label{eq:syn_gamma}
\gamma_{\text{syn}}=\mathrm{clip}\!\left(\exp(\beta_{\text{syn}}\psi),\;\gamma_{\min},\;1\right),
\end{equation}
which downweights policy updates from unstable high-sparsity episodes.
The hyperparameter $\beta_{\text{syn}}$ is tuned so that early training suppresses only catastrophic collapses.

\paragraph{Plan-level GRPO with gated updates.}
User preferences are represented by $\mathbf{w}=[w_{\text{rob}},w_{\text{util}},w_{\text{comp}}]^\top\in\Delta^{2}$ and apply only to robustness, utility, and compression.
For each preference, we sample a group of $G$ plans $\{\mathbf{r}^{(g)}\}_{g=1}^G$ and compute per-plan scalar rewards
\begin{equation}
\label{eq:scalar_reward}
    s^{(g)} = w_{\text{rob}}\,n_{\text{rob}}(\mathbf{r}^{(g)}) \;+\; w_{\text{util}}\,n_{\text{util}}(\mathbf{r}^{(g)}) \;+\; w_{\text{comp}}\,J_{\text{comp}}(\mathbf{r}^{(g)}),
\end{equation}
followed by group-relative advantages
\begin{equation}
\label{eq:grpo_adv}
    \hat{A}^{(g)} = s^{(g)} - \frac{1}{G}\sum_{g'=1}^G s^{(g')}.
\end{equation}
The gated policy-gradient objective under plan-level GRPO~\cite{mroueh2025revisiting} is written in the standard expectation form:
\begin{equation}
\label{eq:actor_loss_expectation}
\begin{aligned}
\mathcal{L}(\phi)
&= -\,\mathbb{E}_{\mathbf{w}\sim p(\mathbf{w}),\,\mathbf{a}^{(1:G)}\sim \pi_\phi(\cdot\mid \mathbf{S},\mathbf{w})}
\left[
\gamma_{\text{syn}} \cdot \frac{1}{G}\sum_{g=1}^{G}
\log \pi_\phi(\mathbf{a}^{(g)}\mid \mathbf{S},\mathbf{w})\;\hat{A}^{(g)}
\right] \\
&\quad - \lambda_{\mathcal{H}}\,\mathbb{E}_{\mathbf{w}\sim p(\mathbf{w})}\!\left[\mathcal{H}\!\left(\pi_\phi(\cdot\mid \mathbf{S},\mathbf{w})\right)\right].
\end{aligned}
\end{equation}
Conceptually, $\gamma_{\text{syn}}$ acts as a \emph{structural trust-region} mechanism: it downweights gradient signals from non-viable pruning episodes, stabilizing the combinatorial plan search under high sparsity.

\subsection{Post-pruning Recovery Fine-tuning}
\label{sec:recovery_ft}

Structured pruning can shift a VLM's output distribution and noticeably degrade hallucination-oriented evaluation, sometimes inducing degenerate Yes/No behavior on POPE.
To account for this practical failure mode, we apply a lightweight \emph{post-pruning recovery} stage that updates a small parameter subset while keeping the structured sparsity mask fixed.
Unless otherwise stated, all pruning plans are compared under a matched recovery budget.

\paragraph{Parameter-efficient adaptation.}
Recovery adapts only a restricted set of parameters, including the LM head and final normalization layers, the multimodal projector, and optionally LoRA adapters on the last $N$ transformer layers~\cite{hu2022lora}.
The pruning mask induced by $\mathbf{r}$ remains unchanged throughout recovery.

\paragraph{POPE-aware stabilization and data.}
Recovery optimizes the standard next-token loss with optional POPE-oriented stabilization to prevent Yes/No collapse: class-balanced token loss, a margin term encouraging $\log p_\theta(y^\star)>\log p_\theta(\bar{y}^\star)$, and a batch-level Yes-rate regularizer.
Recovery is performed on POPE, ScienceQA, or a mixture; when mixed, the sampling ratio can be biased toward POPE (and optionally adjusted according to $\mathbf{w}$) to align post-pruning adaptation with the desired robustness--utility trade-off.

\section{Experiments}
\label{sec:experiments}

\subsection{Experimental Setup}
\label{sec:exp_setup}

\paragraph{Backbones and scope.} 
We evaluate \textbf{HiPP-Prune} on LLaVA-1.5-7B~\cite{liu2023visual} and Qwen2.5-VL-3B~\cite{bai2025qwen25vltechnicalreport}. Pruning targets the language backbone's MLP projections via structured row-wise reduction. All data-aware baselines, including HiPP-Prune, utilize Wanda-style activation statistics~\cite{sun2023simple} estimated on a matched calibration set for fair comparison.

\paragraph{Benchmarks and metrics.} 
We assess two critical post-pruning behaviors: \emph{hallucination robustness} via POPE balanced accuracy (BalAcc)~\cite{li2023evaluating} and \emph{general utility} via ScienceQA (SQA) accuracy~\cite{lu2022learn}. Results are reported alongside the realized structured sparsity of the language backbone.

\paragraph{Evaluation protocol.} 
All methods are compared under matched sparsity budgets (e.g., $\sim$22.5\%). For HiPP-Prune, operating points are obtained by querying the preference-conditioned policy with fixed vectors $\mathbf{w}$. To ensure a controlled probe of initialization quality, all methods undergo an identical post-pruning recovery phase (e.g., 800 steps of LoRA) while keeping the sparsity mask fixed.

\paragraph{Training configuration.} 
HiPP-Prune is optimized using plan-level GRPO~\cite{mroueh2025revisiting}, where candidate blueprints are evaluated in groups to compute relative advantages. Training preferences are sampled from a mixture of discrete anchors and Dirichlet distributions to ensure Pareto coverage. We use a mixed POPE/Utility training set with balanced sampling ratios. Detailed hyperparameters are provided in the supplementary material.

\paragraph{Baselines.}
We compare against representative pruning baselines under matched evaluation settings, including:
(i) \textbf{Random} layer allocation,
(ii) \textbf{Wanda}~\cite{sun2023simple},
(iii) \textbf{SliceGPT}~\cite{ashkboos2024slicegpt}, and
(iv) \textbf{LLM-Pruner}~\cite{ma2023llm}.
Whenever applicable, baseline models are followed by the same recovery protocol as HiPP-Prune for fair comparison.

\subsection{Main Results}
\label{sec:main_results}

We report main results under matched structured sparsity budgets and evaluate three metrics: POPE balanced accuracy (POPE BalAcc), ScienceQA accuracy (SQA Acc), and realized sparsity.
For each baseline, POPE BalAcc and SQA Acc are reported as mean$\pm$std over multiple runs; for HiPP-Prune, we report mean$\pm$std across queried preference points under the same post-pruning recovery protocol.

Table~\ref{tab:main_results_llava7b} summarizes results on LLaVA-7B at sparsity $\approx 0.225$.
Random layer allocations exhibit a large variance, indicating that \emph{where} sparsity is placed can substantially affect robustness and utility even under identical overall sparsity.
Among heuristic baselines, SliceGPT provides a stronger utility point than Random/Wanda, while Wanda does not consistently improve robustness relative to Random at this sparsity level.
In contrast, HiPP-Prune achieves substantially higher POPE BalAcc and higher SQA Acc than all baselines, suggesting that preference-conditioned plan search can discover higher-quality structured pruning plans under the same budget.

Table~\ref{tab:main_results_qwen} further reports results on Qwen2.5-VL-3B under two sparsity budgets ($\approx 0.225$ and $\approx 0.325$).
Across both budgets, HiPP-Prune improves the robustness--utility operating point over heuristic baselines, and the gap remains visible as sparsity increases, indicating that the learned policy continues to produce competitive plans under tighter compression.

\begin{table}[tb]
  \caption{\textbf{Main results on LLaVA-7B at matched sparsity (~22.5\%).} 
  POPE BalAcc and SQA Acc are reported as mean$\pm$std. 
  For HiPP-Prune, we report the average across queried preference points. }
  \label{tab:main_results_llava7b}
  \centering
  \small
  \begin{tabular}{@{}l c cc @{}}
    \toprule
    Method & Sparsity $\downarrow$ & POPE BalAcc (\%) $\uparrow$ & SQA Acc (\%) $\uparrow$ \\
    \midrule
    Dense (Unpruned) & 0.000 & 82.43 & 64.00 \\
    \midrule
    Random & 0.225 & 55.38 $\pm$ 1.99 & 32.83 $\pm$ 0.24 \\
    Wanda & 0.225 & 51.14 $\pm$ 0.20 & 35.75 $\pm$ 1.25 \\
    LLM-Pruner & 0.215 & 49.27 $\pm$ 1.06 & 35.50 $\pm$ 1.00 \\
    SliceGPT & 0.229 & \underline{52.56 $\pm$ 0.04} & \underline{37.75 $\pm$ 0.25} \\
    \rowcolor{gray!10} HiPP-Prune (Ours) & 0.225 & \textbf{72.89 $\pm$ 0.88} & \textbf{39.38 $\pm$ 0.74} \\
    \bottomrule
  \end{tabular}
\end{table}

\begin{table}[tb]
  \caption{\textbf{Main results on Qwen2.5-VL-3B under two sparsity budgets.} 
  POPE BalAcc and SQA Acc are reported as mean$\pm$std over multiple runs; for HiPP-Prune, we report mean$\pm$std across queried preference points. All metrics are in percentages (\%).}
  \label{tab:main_results_qwen}
  \centering
  \footnotesize
  \setlength{\tabcolsep}{5pt} 
  \begin{tabular}{@{}l cc @{\hskip 15pt} cc @{}} 
    \toprule
    \multirow{2}{*}{Method} &
    \multicolumn{2}{c}{\textbf{Sparsity $\approx 0.225$}} &
    \multicolumn{2}{c}{\textbf{Sparsity $\approx 0.325$}} \\
    \cmidrule(lr){2-3} \cmidrule(lr){4-5}
    & POPE  (\%) $\uparrow$ & SQA (\%)$\uparrow$ 
    & POPE  (\%)$\uparrow$ & SQA  (\%)$\uparrow$ \\
    \midrule
    Dense (Unpruned) & 87.36 & 76.54 & 87.36 & 75.42 \\
    \midrule
    Random     & 52.77 $\pm$ 0.19 & 33.81 $\pm$ 3.06
               & 50.86 $\pm$ 2.26 & 31.67 $\pm$ 2.10 \\
    Wanda      & 50.02 $\pm$ 1.98 & 34.36 $\pm$ 0.28
               & 50.02 $\pm$ 1.98 & 29.73 $\pm$ 3.47 \\
    LLM-Pruner & 53.69 $\pm$ 1.03 & 35.79 $\pm$ 0.43
               & 50.39 $\pm$ 1.33 & 26.79 $\pm$ 0.57 \\
    SliceGPT   & \underline{57.17 $\pm$ 0.19} & \underline{36.15 $\pm$ 0.89}
               & \underline{51.17 $\pm$ 0.19} & \underline{31.15 $\pm$ 0.11} \\
    \rowcolor{gray!10} HiPP-Prune(Ours)
               & \textbf{61.65 $\pm$ 3.03} & \textbf{38.91 $\pm$ 0.83}
               & \textbf{53.14 $\pm$ 1.37} & \textbf{32.94 $\pm$ 0.88} \\
    \bottomrule
  \end{tabular}
\end{table}

\subsection{Ablation Study}
\subsubsection{Preference controllability of a single agent}
\label{sec:ablation_pref_query}

To evaluate the zero-shot controllability of our policy, we fix a trained \texttt{best\_agent} and query it with varying preference vectors $\mathbf{w} = [w_{\text{rob}}, w_{\text{util}}, w_{\text{comp}}]$. 
Table~\ref{tab:ablation_pref_query} demonstrates that the hierarchical policy successfully navigates the multi-objective space: 
(i) increasing $w_{\text{rob}}$ boosts POPE performance; 
(ii) higher $w_{\text{util}}$ shifts the allocation to favor ScienceQA; and 
(iii) a larger $w_{\text{comp}}$ prompts the policy to output plans with higher realized sparsity. 
This confirms that a single agent can generalize across the entire Pareto front without retraining.

\begin{table}[tb]
  \caption{\textbf{Preference querying ablation on LLaVA-1.5-7B.} 
  Results are obtained by querying a single fixed policy with different preference weights. 
  Sparsity is the realized structured ratio. POPE and SQA scores are reported in percentages (\%). 
  \textbf{Bold} indicates the highest score in each metric column.}
  \label{tab:ablation_pref_query}
  \centering
  \small
  \begin{tabular}{@{}lccc@{}}
    \toprule
    Preference $\mathbf{w}$ & Sparsity $\downarrow$ & POPE BalAcc (\%) $\uparrow$ & SQA Acc (\%) $\uparrow$ \\
    \midrule
    $(0.60, 0.30, 0.10)$ & 0.2254 & \textbf{74.39} & 39.00 \\
    $(0.45, 0.45, 0.10)$ & 0.2254 & 72.56 & 39.50 \\
    $(0.33, 0.33, 0.34)$ & \textbf{0.2285} & 72.12 & 38.50 \\
    $(0.30, 0.60, 0.10)$ & 0.2253 & 72.48 & \textbf{40.50} \\
    \midrule
    \rowcolor{gray!5} mean $\pm$ std & 0.2262 $\pm$ 0.0014 & 72.89 $\pm$ 0.88 & 39.38 $\pm$ 0.74 \\ 
    \bottomrule
  \end{tabular}
\end{table}

\subsubsection{Preference controllability in operating-point space}
\label{sec:ablation_pref_cloud}

\begin{figure}[!t]
    \centering
    \includegraphics[width=0.98\textwidth]{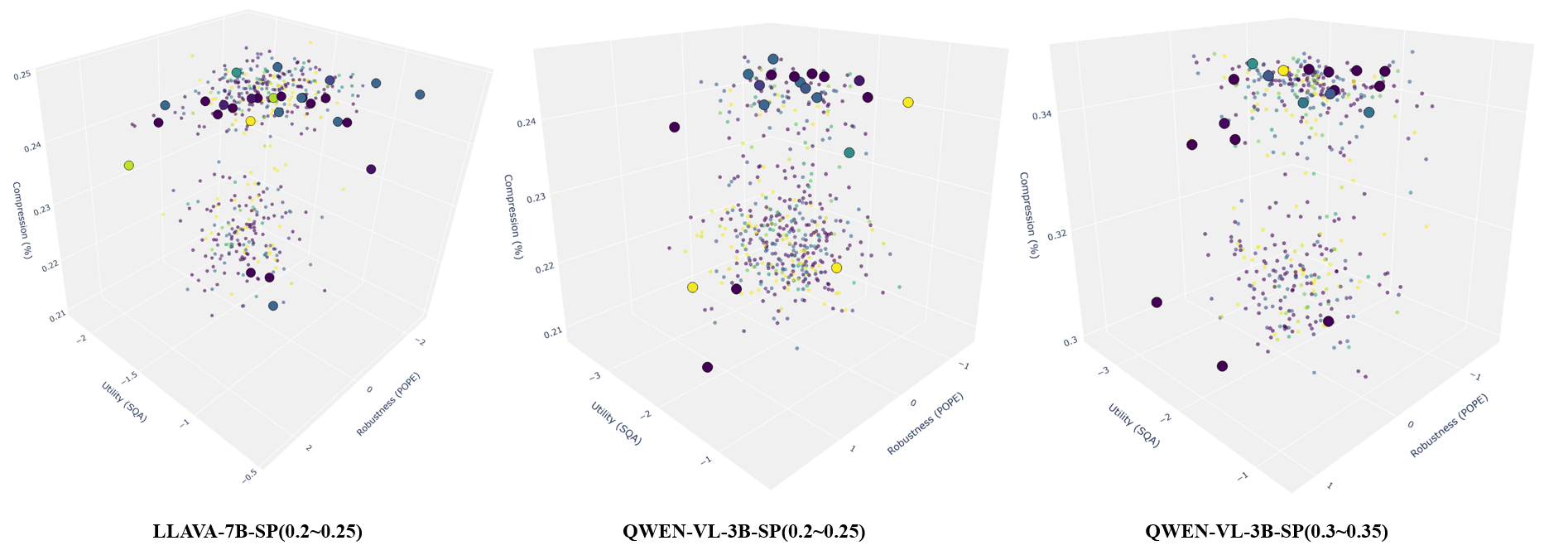}
    \caption{\textbf{Preference-conditioned operating-point clouds and Pareto navigation.} 
    Each point represents a pruning plan generated by a single trained HiPP-Prune policy under varying preference vectors $\mathbf{w}$. 
    Small dots illustrate the breadth of the search space by showing candidate plans sampled during the group-relative optimization process. 
    Large dots denote the final Pareto-optimal blueprints selected by the policy for representative preference queries. 
    Coloring reflects the sampled robustness weight $w_{\text{rob}}$ (ranging from 0 to 1), confirming that the policy effectively navigates the trade-off space between hallucination robustness and task utility by modulating the input vector $\mathbf{w}$.}
    \label{fig:pareto_cloud}
\end{figure}

To evaluate whether a single trained policy can effectively navigate diverse operating points, we visualize the preference-conditioned plan cloud in the robustness--utility--compression space. 
Each point is generated through an \emph{on-the-fly} query to the same HiPP-Prune agent under varying preference vectors $\mathbf{w}$. 
As shown in Fig.~\ref{fig:pareto_cloud}, across different backbones and compression ranges, the resulting plans remain strictly concentrated within the prescribed budget intervals while spanning a wide robustness--utility trade-off region. 
The emergence of a distinct Pareto frontier (large dots) confirms that HiPP-Prune learns a highly controllable policy that avoids solution collapse and instead produces optimal, intent-aligned blueprints under matched budget constraints.

\subsubsection{Preference sampling strategy (anchors vs.\ Dirichlet)}
\label{sec:ablation_sampling}

This ablation investigates how the training-time preference distribution governs the restoration of robustness and utility. We isolate the impact of three sampling strategies: \textbf{Full} (hybrid), \textbf{w/o anchor} (Dirichlet-only), and \textbf{w/o Dirichlet} (anchor-only). All variants are trained for a fixed duration of 3,000 epochs to ensure a controlled comparison.

\paragraph{Matched recovery protocol.}
All pruned models are evaluated \emph{after} the same post-pruning recovery fine-tuning procedure (same trainable scope, data mixture, and optimization budget), so recovered robustness/utility primarily reflect plan quality under fixed compression.

\begin{table}[tb]
  \caption{\textbf{Preference sampling ablation under matched post-pruning recovery.}
  Aggregated over four three-objective preference queries (balanced / robust-heavy / utility-heavy / compression-heavy).}
  \label{tab:ablation_sampling}
  \centering
  \small
  \setlength{\tabcolsep}{6pt}
  \begin{tabular}{@{}lcc@{}}
    \toprule
    Training sampling & POPE (mean$\pm$std) $\uparrow$ & SQA (mean$\pm$std) $\uparrow$ \\
    \midrule
    Full (anchor + Dirichlet)    & 76.62$\pm$0.23 & 36.5$\pm$0.42 \\
    w/o anchor (Dirichlet only)  & 76.01$\pm$1.42 & 36.0$\pm$0.58 \\
    w/o Dirichlet (anchor only)  & 76.34$\pm$0.51 & 36.6$\pm$1.04 \\
    \bottomrule
  \end{tabular}
\end{table}

Under the same recovery budget, the mixed strategy (\textbf{Full}) yields the most stable robustness across queried preferences, with substantially reduced variance compared to Dirichlet-only.
Anchor-only attains slightly higher mean utility, consistent with specializing to a small set of discrete anchor preferences, whereas Dirichlet-only exhibits larger sensitivity to preference changes.
Overall, combining anchor sampling (as boundary supervision) with Dirichlet sampling (as interior coverage) provides a favorable balance between robustness stability and utility in this 3000-epoch training regime.

\section{Conclusion}
\label{sec:conclusion}

This paper introduced \textbf{HiPP-Prune}, a hierarchical preference-conditioned framework that reformulates VLM structured pruning as a conditional resource allocation task. By learning a plan-level policy integrated with attention-flow-based visual sensitivity and a SynFlow-inspired stability gate, our approach generates diverse layer-wise blueprints that protect vision-critical components even under aggressive compression.

Evaluations on LLaVA and Qwen2.5-VL demonstrate that a single HiPP-Prune agent can effectively navigate robustness--utility trade-offs via zero-shot preference querying, providing a flexible ``query-once'' mechanism for varied deployment constraints. Empirical results show that our learned plans consistently yield pruned initializations that recover to superior performance compared to heuristic baselines under matched recovery budgets. While a performance gap persists at extreme sparsity levels, our findings highlight the potential of adaptive allocation as a first-class consideration in multimodal compression.

\bibliographystyle{unsrtnat}
\bibliography{main}  

@String(ICLR  = {Int. Conf. Learn. Represent.})

@String(AAAI  = {AAAI})

@String(ICLR  = {ICLR})

@article{liu2023visual,
  title={Visual instruction tuning},
  author={Liu, Haotian and Li, Chunyuan and Wu, Qingyang and Lee, Yong Jae},
  journal={Advances in neural information processing systems},
  volume={36},
  pages={34892--34916},
  year={2023}
}

@inproceedings{li2023evaluating,
  title={Evaluating object hallucination in large vision-language models},
  author={Li, Yifan and Du, Yifan and Zhou, Kun and Wang, Jinpeng and Zhao, Wayne Xin and Wen, Ji-Rong},
  booktitle={Proceedings of the 2023 conference on empirical methods in natural language processing},
  pages={292--305},
  year={2023}
}

@inproceedings{leng2024mitigating,
  title={Mitigating object hallucinations in large vision-language models through visual contrastive decoding},
  author={Leng, Sicong and Zhang, Hang and Chen, Guanzheng and Li, Xin and Lu, Shijian and Miao, Chunyan and Bing, Lidong},
  booktitle={Proceedings of the IEEE/CVF Conference on Computer Vision and Pattern Recognition},
  pages={13872--13882},
  year={2024}
}

@article{han2015learning,
  title={Learning both weights and connections for efficient neural network},
  author={Han, Song and Pool, Jeff and Tran, John and Dally, William},
  journal={Advances in neural information processing systems},
  volume={28},
  year={2015}
}

@article{frankle2018lottery,
  title={The lottery ticket hypothesis: Finding sparse, trainable neural networks},
  author={Frankle, Jonathan and Carbin, Michael},
  journal={arXiv preprint arXiv:1803.03635},
  year={2018}
}

@article{sun2023simple,
  title={A simple and effective pruning approach for large language models},
  author={Sun, Mingjie and Liu, Zhuang and Bair, Anna and Kolter, J Zico},
  journal={arXiv preprint arXiv:2306.11695},
  year={2023}
}

@inproceedings{frantar2023sparsegpt,
  title={Sparsegpt: Massive language models can be accurately pruned in one-shot},
  author={Frantar, Elias and Alistarh, Dan},
  booktitle={International conference on machine learning},
  pages={10323--10337},
  year={2023},
  organization={PMLR}
}

@article{roijers2013survey,
  title={A survey of multi-objective sequential decision-making},
  author={Roijers, Diederik M and Vamplew, Peter and Whiteson, Shimon and Dazeley, Richard},
  journal={Journal of Artificial Intelligence Research},
  volume={48},
  pages={67--113},
  year={2013}
}

@inproceedings{kwon2023efficient,
  title={Efficient Memory Management for Large Language Model Serving with PagedAttention},
  author={Woosuk Kwon and Zhuohan Li and Siyuan Zhuang and Ying Sheng and Lianmin Zheng and Cody Hao Yu and Joseph E. Gonzalez and Hao Zhang and Ion Stoica},
  booktitle={Proceedings of the ACM SIGOPS 29th Symposium on Operating Systems Principles},
  year={2023}
}

@article{lu2022learn,
  title={Learn to explain: Multimodal reasoning via thought chains for science question answering},
  author={Lu, Pan and Mishra, Swaroop and Xia, Tanglin and Qiu, Liang and Chang, Kai-Wei and Zhu, Song-Chun and Tafjord, Oyvind and Clark, Peter and Kalyan, Ashwin},
  journal={Advances in neural information processing systems},
  volume={35},
  pages={2507--2521},
  year={2022}
}

@article{mroueh2025revisiting,
  title={Revisiting group relative policy optimization: Insights into on-policy and off-policy training},
  author={Mroueh, Youssef and Dupuis, Nicolas and Belgodere, Brian and Nitsure, Apoorva and Rigotti, Mattia and Greenewald, Kristjan and Navratil, Jiri and Ross, Jerret and Rios, Jesus},
  journal={arXiv preprint arXiv:2505.22257},
  year={2025}
}

@inproceedings{liu2025efficient,
  title={Efficient discovery of Pareto front for multi-objective reinforcement learning},
  author={Liu, Ruohong and Pan, Yuxin and Xu, Linjie and Song, Lei and You, Pengcheng and Chen, Yize and Bian, Jiang},
  booktitle={The Thirteenth International Conference on Learning Representations},
  year={2025}
}

@inproceedings{huang2024ivtp,
  title={Ivtp: Instruction-guided visual token pruning for large vision-language models},
  author={Huang, Kai and Zou, Hao and Xi, Ye and Wang, BoChen and Xie, Zhen and Yu, Liang},
  booktitle={European conference on computer vision},
  pages={214--230},
  year={2024},
  organization={Springer}
}

@inproceedings{ye2025atp,
  title={Atp-llava: Adaptive token pruning for large vision language models},
  author={Ye, Xubing and Gan, Yukang and Ge, Yixiao and Zhang, Xiao-Ping and Tang, Yansong},
  booktitle={Proceedings of the IEEE/CVF Conference on Computer Vision and Pattern Recognition},
  pages={24972--24982},
  year={2025}
}

@inproceedings{ma2025short,
  title={Short-lvlm: Compressing and accelerating large vision-language models by pruning redundant layers},
  author={Ma, Ji and Suo, Wei and Wang, Peng and Zhang, Yanning},
  booktitle={Proceedings of the 33rd ACM International Conference on Multimedia},
  pages={3575--3584},
  year={2025}
}

@article{he2025rethinking,
  title={Rethinking Pruning for Vision-Language Models: Strategies for Effective Sparsity},
  author={He, Shwai and Li, Ang and Chen, Tianlong},
  journal={ACM SIGMETRICS Performance Evaluation Review},
  volume={53},
  number={2},
  pages={9--14},
  year={2025},
  publisher={ACM New York, NY, USA}
}

@inproceedings{liang2025efficientllava,
  title={Efficientllava: generalizable auto-pruning for large vision-language models},
  author={Liang, Yinan and Wang, Ziwei and Xu, Xiuwei and Zhou, Jie and Lu, Jiwen},
  booktitle={Proceedings of the Computer Vision and Pattern Recognition Conference},
  pages={9445--9454},
  year={2025}
}

@article{ma2023llm,
  title={Llm-pruner: On the structural pruning of large language models},
  author={Ma, Xinyin and Fang, Gongfan and Wang, Xinchao},
  journal={Advances in neural information processing systems},
  volume={36},
  pages={21702--21720},
  year={2023}
}

@article{hu2022lora,
  title={Lora: Low-rank adaptation of large language models.},
  author={Hu, Edward J and Shen, Yelong and Wallis, Phillip and Allen-Zhu, Zeyuan and Li, Yuanzhi and Wang, Shean and Wang, Liang and Chen, Weizhu and others},
  journal={Iclr},
  volume={1},
  number={2},
  pages={3},
  year={2022}
}

@article{seale2025flexigpt,
  title={FlexiGPT: Pruning and Extending Large Language Models with Low-Rank Weight Sharing},
  author={Seale Smith, James and Lin, Chi-Heng and Tuli, Shikhar and Jeelani, Haris and Gao, Shangqian and Shen, Yilin and Jin, Hongxia and Hsu, Yen-Chang},
  journal={arXiv e-prints},
  pages={arXiv--2501},
  year={2025}
}

@article{tanaka2020pruning,
  title={Pruning neural networks without any data by iteratively conserving synaptic flow},
  author={Tanaka, Hidenori and Kunin, Daniel and Yamins, Daniel L and Ganguli, Surya},
  journal={Advances in neural information processing systems},
  volume={33},
  pages={6377--6389},
  year={2020}
}

@article{lee2018snip,
  title={Snip: Single-shot network pruning based on connection sensitivity},
  author={Lee, Namhoon and Ajanthan, Thalaiyasingam and Torr, Philip HS},
  journal={arXiv preprint arXiv:1810.02340},
  year={2018}
}

@article{wang2020picking,
  title={Picking winning tickets before training by preserving gradient flow},
  author={Wang, Chaoqi and Zhang, Guodong and Grosse, Roger},
  journal={arXiv preprint arXiv:2002.07376},
  year={2020}
}

@article{schulman2017proximal,
  title={Proximal policy optimization algorithms},
  author={Schulman, John and Wolski, Filip and Dhariwal, Prafulla and Radford, Alec and Klimov, Oleg},
  journal={arXiv preprint arXiv:1707.06347},
  year={2017}
}

@InProceedings{pmlr-v37-schulman15,
  title = 	 {Trust Region Policy Optimization},
  author = 	 {Schulman, John and Levine, Sergey and Abbeel, Pieter and Jordan, Michael and Moritz, Philipp},
  booktitle = 	 {Proceedings of the 32nd International Conference on Machine Learning},
  pages = 	 {1889--1897},
  year = 	 {2015},
  editor = 	 {Bach, Francis and Blei, David},
  volume = 	 {37},
  series = 	 {Proceedings of Machine Learning Research},
  address = 	 {Lille, France},
  month = 	 {07--09 Jul},
  publisher =    {PMLR},
  url = 	 {https://proceedings.mlr.press/v37/schulman15.html}
}

@InProceedings{pmlr-v70-achiam17a,
  title = 	 {Constrained Policy Optimization},
  author =       {Joshua Achiam and David Held and Aviv Tamar and Pieter Abbeel},
  booktitle = 	 {Proceedings of the 34th International Conference on Machine Learning},
  pages = 	 {22--31},
  year = 	 {2017},
  editor = 	 {Precup, Doina and Teh, Yee Whye},
  volume = 	 {70},
  series = 	 {Proceedings of Machine Learning Research},
  month = 	 {06--11 Aug},
  publisher =    {PMLR},
  url = 	 {https://proceedings.mlr.press/v70/achiam17a.html}
}

@article{ashkboos2024slicegpt,
  title={Slicegpt: Compress large language models by deleting rows and columns},
  author={Ashkboos, Saleh and Croci, Maximilian L and Nascimento, Marcelo Gennari do and Hoefler, Torsten and Hensman, James},
  journal={arXiv preprint arXiv:2401.15024},
  year={2024}
}

@misc{bai2025qwen25vltechnicalreport,
      title={Qwen2.5-VL Technical Report}, 
      author={Shuai Bai and Keqin Chen and Xuejing Liu and Jialin Wang and Wenbin Ge and Sibo Song and Kai Dang and Peng Wang and Shijie Wang and Jun Tang and Humen Zhong and Yuanzhi Zhu and Mingkun Yang and Zhaohai Li and Jianqiang Wan and Pengfei Wang and Wei Ding and Zheren Fu and Yiheng Xu and Jiabo Ye and Xi Zhang and Tianbao Xie and Zesen Cheng and Hang Zhang and Zhibo Yang and Haiyang Xu and Junyang Lin},
      year={2025},
      eprint={2502.13923},
      archivePrefix={arXiv},
      primaryClass={cs.CV},
      url={https://arxiv.org/abs/2502.13923}, 
}

@inproceedings{guan2024hallusionbench,
  title={Hallusionbench: an advanced diagnostic suite for entangled language hallucination and visual illusion in large vision-language models},
  author={Guan, Tianrui and Liu, Fuxiao and Wu, Xiyang and Xian, Ruiqi and Li, Zongxia and Liu, Xiaoyu and Wang, Xijun and Chen, Lichang and Huang, Furong and Yacoob, Yaser and others},
  booktitle={Proceedings of the IEEE/CVF conference on computer vision and pattern recognition},
  pages={14375--14385},
  year={2024}
}

@inproceedings{alvar2025divprune,
  title={Divprune: Diversity-based visual token pruning for large multimodal models},
  author={Alvar, Saeed Ranjbar and Singh, Gursimran and Akbari, Mohammad and Zhang, Yong},
  booktitle={Proceedings of the Computer Vision and Pattern Recognition Conference},
  pages={9392--9401},
  year={2025}
}

@inproceedings{ye2025fit,
  title={Fit and prune: Fast and training-free visual token pruning for multi-modal large language models},
  author={Ye, Weihao and Wu, Qiong and Lin, Wenhao and Zhou, Yiyi},
  booktitle={Proceedings of the AAAI Conference on Artificial Intelligence},
  volume={39},
  pages={22128--22136},
  year={2025}
}

@inproceedings{zhuang2025st3,
  title={St3: Accelerating multimodal large language model by spatial-temporal visual token trimming},
  author={Zhuang, Jiedong and Lu, Lu and Dai, Ming and Hu, Rui and Chen, Jian and Liu, Qiang and Hu, Haoji},
  booktitle={Proceedings of the AAAI Conference on Artificial Intelligence},
  volume={39},
  pages={11049--11057},
  year={2025}
}

@inproceedings{sun2025lvpruning,
  title={Lvpruning: An effective yet simple language-guided vision token pruning approach for multi-modal large language models},
  author={Sun, Yizheng and Xin, Yanze and Li, Hao and Sun, Jingyuan and Lin, Chenghua and Batista-Navarro, Riza Theresa},
  booktitle={Findings of the Association for Computational Linguistics: NAACL 2025},
  pages={4299--4308},
  year={2025}
}

@inproceedings{huang2025prunevid,
  title={Prunevid: Visual token pruning for efficient video large language models},
  author={Huang, Xiaohu and Zhou, Hao and Han, Kai},
  booktitle={Findings of the Association for Computational Linguistics: ACL 2025},
  pages={19959--19973},
  year={2025}
}






\end{document}